%% file: main.tex
\documentclass[10pt,twocolumn,letterpaper]{article}

\usepackage[pagenumbers]{iccv} 


\definecolor{iccvblue}{rgb}{0.21,0.49,0.74}
\usepackage[pagebackref,breaklinks,colorlinks,allcolors=iccvblue]{hyperref}
\usepackage{colortbl}
\usepackage[ruled]{algorithm2e}
\newcommand{\dn}{\texttt{OmniRig}}
\def\paperID{4387} 
\def\confName{ICCV}
\def\confYear{2025}

\title{ARMO: Autoregressive Rigging for Multi-Category Objects}

\def\authorBlock{
    Mingze Sun\textsuperscript{1}\textsuperscript{*} \quad 
    Shiwei Mao\textsuperscript{1}\textsuperscript{*} \quad
    Keyi Chen\textsuperscript{1} \quad
    Yurun Chen\textsuperscript{1} \quad
    Shunlin Lu\textsuperscript{2} \quad
    Jingbo Wang\textsuperscript{3} \quad
    Junting Dong\textsuperscript{3}\textsuperscript{\textdagger}  \quad
    Ruqi Huang\textsuperscript{1}\textsuperscript{\textdagger} \\
    
    \textsuperscript{1}Tsinghua Shenzhen International Graduate School, China \\
    \textsuperscript{2}The Chinese University of Hong Kong, Shenzhen
    \\  
    \textsuperscript{3}Shanghai AI Laboratory, China \\
}

\begin{document}

\twocolumn[{

\renewcommand\twocolumn[1][]{#1}
\maketitle

\begin{center}
    \vspace{-4\baselineskip}
    \author{\authorBlock}
\end{center}

\begin{center}
    \captionsetup{type=figure}
    \centerline{\includegraphics[width=\linewidth]{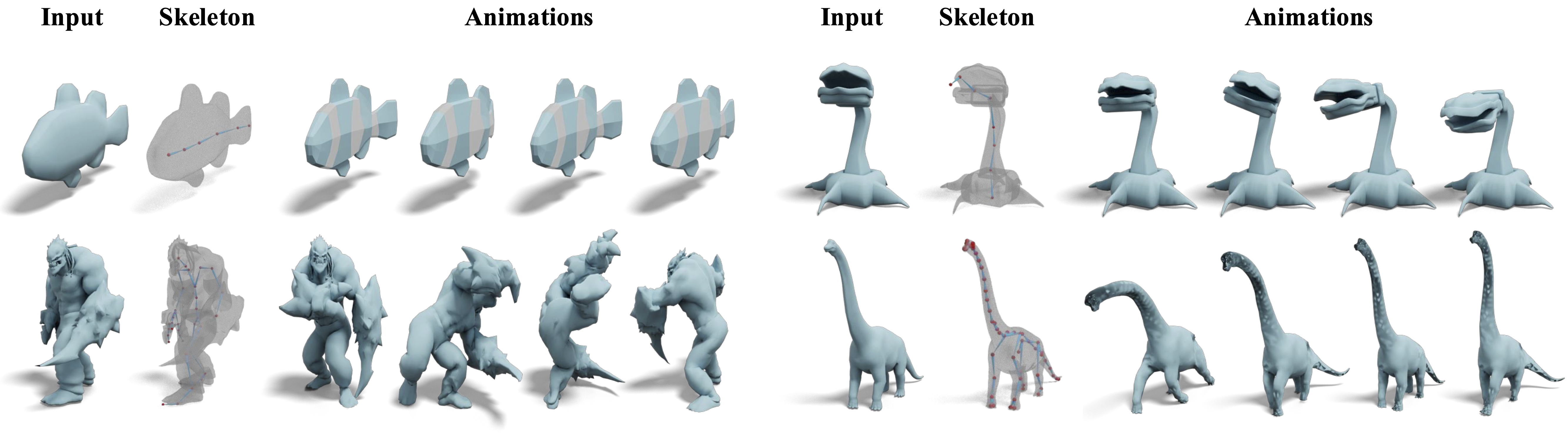}}
    \caption{
    We propose \textbf{ARMO}, a pipeline that generates accurate skeleton results from mesh inputs, enabling precise control over the mesh to generate realistic and accurately driven results.
    }
    \label{fig:teaser}

\end{center}
}]

\renewcommand{\thefootnote}{} 
\footnotetext{\textsuperscript{*} Indicates Equal Contribution. \textsuperscript{\textdagger} Indicates Corresponding Author.}
\renewcommand{\thefootnote}{\arabic{footnote}} 

\input{sec/0_abstract}    
\input{sec/1_intro}
\input{sec/2_related}

\input{sec/3_dataset}
\input{sec/4_method}
\input{sec/5_exp}
\input{sec/6_app}
\input{sec/7_conclusion}
{
    \small
    \bibliographystyle{ieeenat_fullname}
    \bibliography{main}
}

\end{document}

%% file: sec/0_abstract.tex
\begin{abstract}

Recent advancements in large-scale generative models have significantly improved the quality and diversity of 3D shape generation. However, most existing methods focus primarily on generating static 3D models, overlooking the potential dynamic nature of certain shapes, such as humanoids, animals, and insects. To address this gap, we focus on rigging, a fundamental task in animation that establishes skeletal structures and skinning for 3D models.
In this paper, we introduce $\dn$, the first large-scale rigging dataset, comprising 79,499 meshes with detailed skeleton and skinning information. 
Unlike traditional benchmarks that rely on predefined standard poses (e.g., A-pose, T-pose), our dataset embraces diverse shape categories, styles, and poses. Leveraging this rich dataset, we propose \texttt{ARMO}, a novel rigging framework that utilizes an autoregressive model to predict both joint positions and connectivity relationships in a unified manner. 
By treating the skeletal structure as a complete graph and discretizing it into tokens, we encode the joints using an auto-encoder to obtain a latent embedding and an autoregressive model to predict the tokens
A mesh-conditioned latent diffusion model is used to predict the latent embedding for conditional skeleton generation.
Our method addresses the limitations of regression-based approaches, which often suffer from error accumulation and suboptimal connectivity estimation. 
Through extensive experiments on the $\dn$ dataset, our approach achieves state-of-the-art performance in skeleton prediction, demonstrating improved generalization across diverse object categories. 
The code and dataset will be made public for academic use upon acceptance.

\end{abstract}

%% file: sec/1_intro.tex
\section{Introduction}
Recently, large-scale generative models have achieved impressive advancement on generating 3D shapes of high quality and great diversity through multi-modal hints, such as text~\cite{wang2024llama, liang2024luciddreamer}, images~\cite{wu2024unique3d, zhang2024clay, long2024wonder3d}, or point clouds~\cite{chen2024meshanything, chen2024meshanythingv2}. 
However, the majority of efforts in this field focuses on delivering \emph{static} digitizations, to some extent overlooking the potential dynamic nature of the shapes of interest (\emph{e.g.,} humanoids, animals, insects). 
To alleviate this discrepancy, we resort to rigging, a long-standing task from animation research, which creates a skeletal structure for a 3D shape and further relates shape to the simplified skeleton. 
High-quality rigging allows for driving shapes through skeletal motions crafted by artists and, hopefully, more and more advanced automatic algorithms~\cite{gat2025anytop}.
In this paper, we propose a novel rigging framework, \texttt{ARMO}, for rigging 3D shapes, which can manifest \emph{significant diversity in style, pose, and category}. 

The first challenge that arose in our study is the lack of \emph{large-scale} dataset with high-quality rigging annotations. 
In fact, though automatic rigging algorithms~\cite{xu2020rignet, chu2024humanrig, guo2024make} have long been studied, the regarding dataset construction like Modelresource~\cite{xu2020rignet} and Mixamo~\cite{mixamo} is lagged. 
To this end, we have dedicated to constructing, to the best of our knowledge, the first large-scale rigging dataset, $\dn$, which consists of 79,499 meshes with detailed skeleton and skinning information.
We conducted extensive data cleaning from ObjaverseXL and publicly available websites, followed by manual annotation of the data categories.

The second challenge then follows close on the establishment of $\dn$, namely, how can we fully exploit the potential of the large-scale data? RigNet~\cite{xu2020rignet} and TARig~\cite{ma2023tarig} use mesh as input and design regression networks to predict the rigging results.
Regression-based methods directly predict the full set of joint coordinates and the corresponding probability matrix for connectivity. 

Though these prior works have greatly advanced rigging performance in the past decades, even with extensive training on large datasets, such designs often struggle to generalize well in scenarios involving significant data variability and diverse object categories, making it challenging for them to achieve satisfactory results on our proposed rigging dataset $\dn$, which includes a diverse range of shape categories. 
However, the autoregressive model adopts a next-token prediction strategy, which effectively leverages both the given conditions and previously predicted results for iterative prediction. 
Moreover, autoregressive models have been extensively validated in other domains, demonstrating strong generalization capabilities when trained on large-scale datasets~\cite{siddiqui2024meshgpt, chen2024meshanything, chen2024meshanythingv2}.
To address the problem above, we propose a novel approach focused on accurate skeleton estimation that models the skeletal structure as a complete graph and employs an autoregressive model to learn both the joint positions and their corresponding parent joint positions.

Additionally, these prior works all follow a multi-stage design (\emph{e.g., }formulating independent modules for joint prediction, bone connection, and skinning), which significantly limits their utility in our dataset. 
These approaches are prone to error accumulation, and using greedy algorithms for bone connectivity estimation often leads to suboptimal results.
Errors in bone connectivity can significantly impact both skinning estimation and motion control. 
Our proposed approach, which employs an autoregressive model, offers a more structured way to learn the skeleton. 
By predicting each joint position along with its corresponding parent joint position, our method directly infers bone connectivity, effectively reducing error propagation and improving overall prediction accuracy.

Specifically, we first represent the skeleton as a tree structure, where each joint is associated with the position of its parent joint. 
We then discretize this tree structure into tokens. 
The joints are first encoded using an autoregressive auto-encoder, which produces a latent embedding. 
This embedding is subsequently processed by an autoregressive model to predict the skeletal structure tokens.
Previous studies have shown that directly applying a conditioned autoregressive model can lead to confusion between the conditioning input and the output~\cite{tang2024edgerunner}, resulting in degraded prediction performance.
Building on this, we further train a mesh-conditioned latent diffusion model to predict the latent embedding, enabling accurate conditioned skeleton generation.

Our key contributions are threefold: 1) We introduce $\dn$, the first large-scale rigging dataset, which covers diverse object categories with detailed rigging annotations.
2) We propose, for the first time, an autoregressive model that simultaneously predicts both joint positions and connectivity relationships.
3) We conduct extensive experiments on the $\dn$ dataset, achieving state-of-the-art performance in skeleton prediction.

%% file: sec/2_related.tex
\section{Related Works}

\subsection{3D Autoregressive Models}
Autoregressive models, which are first designed for natural language processing~\cite{vaswani2017attention, radford2019language}, have recently gained rapid development in 2D image processing~\cite{sun2024autoregressive, lee2022autoregressive, tian2025visual} and 3D generative models. The main idea is to automatically predict the next component in a sequence by taking measurements from previous input. MeshGPT~\cite{siddiqui2024meshgpt} represents meshes as latent embeddings through geometry and topology and proposes a sequence-based method to autoregressively generate meshes as a series of triangles. In contrast, MeshXL~\cite{chen2025meshxl} explores the neural coordinate field to construct an explicit representation for 3D meshes and formulate several base models for different 3D mesh generation tasks. For the sake of generating high-quality meshes with geometric features, MeshAnything~\cite{chen2024meshanythingv2} introduces a shape-conditioned auto-regressive transformer to align the generated meshes with given shapes. Based on this, MeshAnything V2~\cite{chen2024meshanything} creates adjacent mesh tokenization, further increasing the efficiency of mesh embedding and the performance of the generated meshes. Instead of focusing on tokenizing the complicated topology of the meshes, PivotMesh~\cite{weng2024pivotmesh} encodes meshes into discrete tokens and realizes a scalable mesh generation framework. However, these tokenization algorithms are still insufficient to handle high-resolution and complex objects. To solve this question, EdgeRunner~\cite{tang2024edgerunner} compresses variable-length meshes into fixed-length latent codes and demonstrates that latent embedding can increase generalization and robustness.

\subsection{Automatic Rigging}
In the field of rigging, the process has traditionally relied heavily on manual effort by designers, making it both time-consuming and labor-intensive. In recent years, the rapid advancement of machine learning techniques has paved the way for automatic rigging methods.
It aims to create reasonable skeletons for 3D assets without manual rigging and utilizes skeletal motion data to animate. 

In automatic rigging, extensive research has been conducted on human rigging. 
A notable milestone in this field is Neural Body~\cite{peng2021neural}, which pioneered the use of the SMPL model for automatically generating dynamic 3D human models. This approach laid the groundwork for various methods that employ the SMPL model for animatable 3D human generation, including ~\cite{hu2023sherf, su2023caphy, xu2024xagen}.
Methods like PointSkelCNN~\cite{qin2020pointskelcnn} and S3~\cite{yang2021s3} aim to learn rigging data from labeled human body skeletons rather than relying directly on the SMPL model. 
In addition, for the sake of creating high-quality skeletons, recent methods~\cite{guo2024make, chu2024humanrig, sun2024drive} choose to focus on heterogeneous humanoid characters and achieve satisfactory performances.
However, these methods are limited to this specific category. 
In broader applications, achieving automatic rigging for arbitrary shapes is becoming increasingly important. 

Existing rigging approaches for arbitrary shapes can generally be categorized into two types: optimization-based methods and learning-based methods. 
Among optimization-based methods, Pinocchio~\cite{baran2007automatic} is a pioneering method in this research area, which adapts a predefined template skeleton to the mesh. 
CASA~\cite{wu2022casa} was the first to propose jointly inferring articulated skeletal structures and rigging parameters through optimization. 
Later developments have integrated techniques such as dynamic NeRF~\cite{yang2022banmo} and dual-phase optimization~\cite{zhang2024s3o} to enhance both 3D object reconstruction and rigging quality.
Despite their effectiveness, optimization-based methods are inherently instance-specific, limiting their generalization ability. Consequently, they are often computationally expensive and impractical for large-scale data processing or for rigging objects with highly diverse structures.

Consequently, Li et al.~\cite{li2021learning} explore the learning-based method and improve the quality of mesh deformation.
TARig~\cite{ma2023tarig} proposes an adaptive template skeleton and introduces a boneflow component to improve the structure of the skeleton. However, these template-based methods are limited to specific categories or standard poses, making them difficult to generalize to diverse objects and topologies. 
In contrast, RigNet~\cite{xu2020rignet} takes a mesh as input and employs dedicated networks to independently predict joint positions, bone connections, and skinning weights.
For joint estimation, RigNet first predicts position offsets using a regression-based approach. It then performs differentiable clustering, where the final joint positions are determined by the cluster centers. 
Subsequently, the Minimum Spanning Tree (MST) algorithm is used to establish connectivity between the unordered joints.
This sequential pipeline introduces additional complexity and is susceptible to error accumulation across stages. 
Furthermore, regression-based predictions often suffer from limited generalization ability, making them less effective when applied to diverse or unseen shapes.
We propose using an autoregressive model to simultaneously predict both joint positions and connectivity relationships, which effectively reduces error accumulation. 
Furthermore, the autoregressive model demonstrates improved generalization performance after being trained on a large-scale dataset.

%% file: sec/3_dataset.tex
\section{Datasets}
To address the persistent challenge of data scarcity in rigging research, we introduce $\dn$, a comprehensive and large-scale dataset with detailed rigging annotations. Our dataset is constructed from three key sources: ModelResource\cite{xu2020rignet}, ObjaverseXL\cite{objaverseXL}, and publicly available free data collected from the internet. In total, $\dn$ comprises 79,499 meshes, each accompanied by detailed rigging information, making it the largest and most diverse rigging dataset to date.

The construction of $\dn$ follows a two-stage process: data filtering and post-processing. During these stages, we employ a combination of manual inspection and automated methods to ensure data quality, diversity, and completeness. Below, we describe each stage in detail.

\subsection{Data filtering}
Our data filtering process is designed to ensure that only high-quality models with valid rigging information are retained. The dataset is constructed from three primary sources: ModelResource, ObjaverseXL, and publicly available online data. ModelResource contains 2,354 high-quality 3D models, each equipped with both skeleton and skinning information, serving as reliable references for human-centric rigging tasks. ObjaverseXL is a large-scale dataset with over 8 million 3D models; from this extensive collection, we selectively extract 300,000 models in FBX and GLB formats, which are widely used in rigging applications. Additionally, to enhance the dataset's diversity, we include 1,100 models collected from freely available online resources, ensuring broader coverage of object categories.

To construct a clean and high-quality dataset, we apply a two-stage filtering process. In the initial filtering stage, we discard models that lack skeleton and skinning information, exhibit poor mesh quality, or contain mismatched skeleton and skinning data. Given the large volume of data in ObjaverseXL, we observe several cases where meshes are misaligned with their corresponding skeletons or contain anatomically unreasonable skeletal structures. To address this, we render both meshes and their corresponding skeletons as images, allowing us to manually inspect and remove models with clearly erroneous skeleton structures. This additional quality assessment step ensures that only meaningful and accurate rigging data is preserved.

Following this rigorous filtering process, we retain 76,045 high-quality models. 
To further diversify the dataset, we integrate the 1,100 additional models from publicly available sources using the same filtering criteria. 
In total, our dataset comprises 79,499 models with rigging information. 
Notably, the dataset features skeletal structures with varying complexity, with the number of joints ranging from 2 to 100, ensuring suitability for a wide range of rigging applications. 
A visualization of part of our dataset, showcasing various high-quality meshes along with their corresponding rigging information, is presented in Fig.~\ref{fig:dataset}
\input{figs/dataset}

\subsection{Post processing}
Following the data filtering process, we perform additional post-processing steps to further organize and structure the dataset for effective use in research tasks. For each filtered model, we extract the mesh along with its associated skeleton and skinning information. 
To facilitate research in category-specific rigging tasks, we manually annotate the dataset with meaningful category labels. Given the inconsistency in humanoid rigging across different sources, we classify the models into eight distinct categories to better capture the dataset's diversity: complex characters (with finger bones), simple characters (without finger bones), animals, marine creatures, birds, insects, plants, and others. 
This categorization not only provides valuable insights into the dataset's composition but also enables targeted research such as category-specific rigging augmentation and skeleton learning. 

A visualization of the dataset’s category distribution is shown in Fig.~\ref{fig:wordcloud}, which highlights the dominance of character data while illustrating the dataset's richness in non-character categories as well. 
We believe that the diverse range of object categories and detailed rigging annotations provided in our dataset will significantly benefit future research in 3D rigging, pose estimation, and animation synthesis, while also serving as a valuable resource for developing more robust and generalizable rigging models.

\input{figs/wordcloud}

%% file: figs/dataset.tex
\begin{figure}[h!]
  \begin{center}
\includegraphics[width=\linewidth]{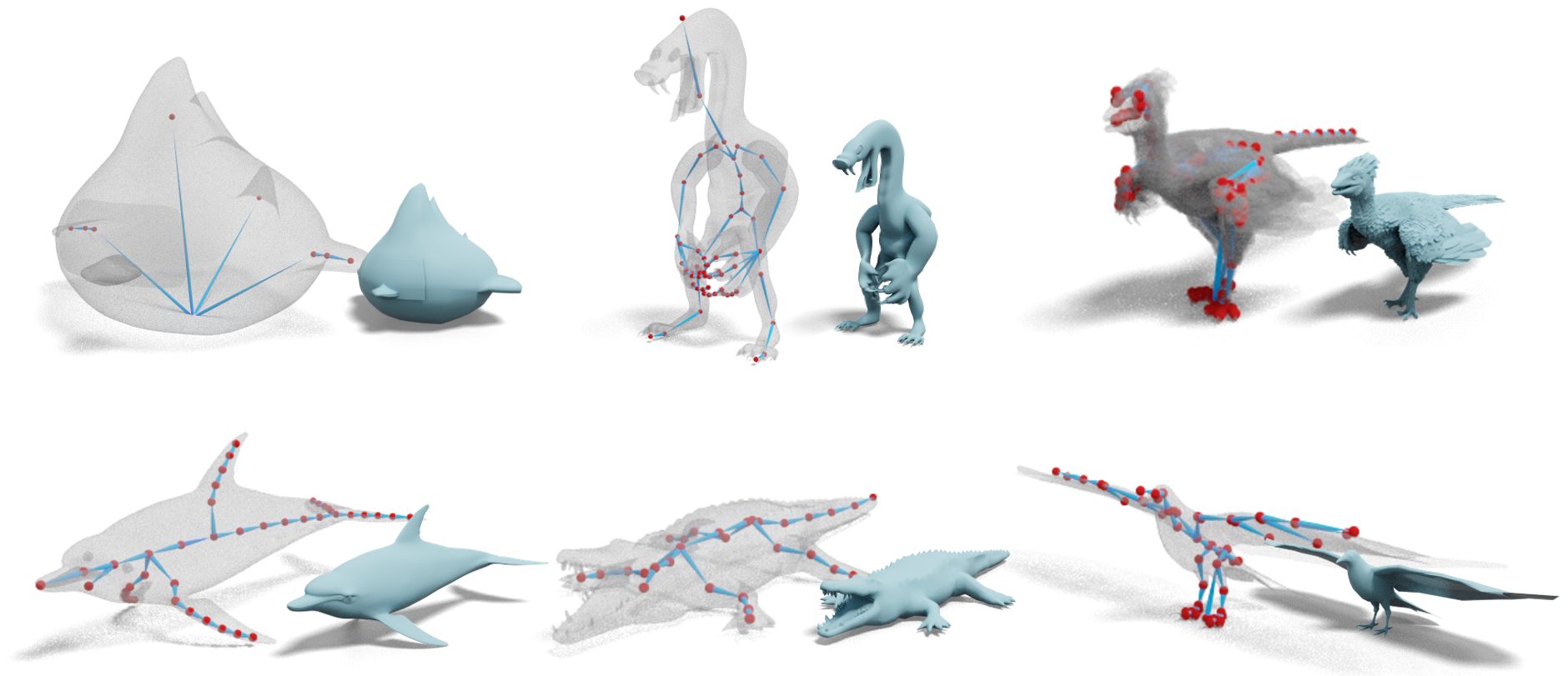}
  \end{center}
\caption{Visualization of the dataset $\dn$, which contains high-quality skeleton structures and objects in diverse categories.}
\label{fig:dataset}
\end{figure}

%% file: figs/wordcloud.tex
\begin{figure}[h!]
  \begin{center}
\includegraphics[width=6.5cm]{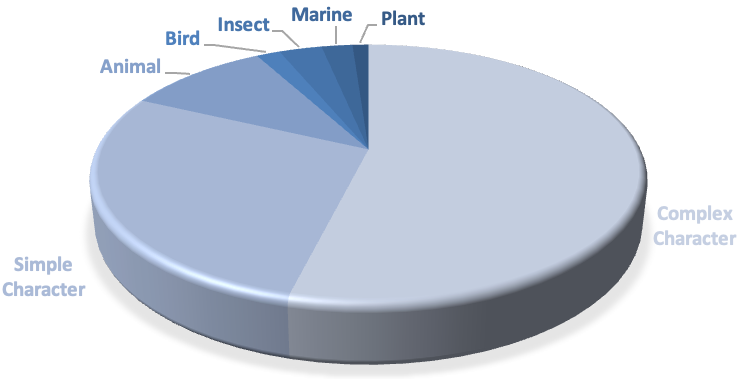}
  \end{center}
\caption{A pie chart indicating the multiple categories in our largr-scale rigging dataset $\dn$.}
\label{fig:wordcloud}
\end{figure}

%% file: sec/4_method.tex
\section{Methodology}
\input{figs/pipeline}
In this section, we present the overall pipeline of our
rigging system, which is trained on our $\dn$ dataset.
Our model focuses on generating high-quality skeletal structures. In Sec.~\ref{sec:problem}, we introduce our problem formulation and provide a brief overview of our autoregressive model. 
In Sec.~\ref{sec:skeleton}, we introduce our autoregressive skeleton generation model, which includes skeleton reconstruction based on an autoregressive auto-encoder and diffusion-based point cloud conditional generation.

\subsection{Problem formulation}\label{sec:problem}
Given a mesh $\mathbf{M}$ with vertices $\mathbf{V}\in \mathbb{R}^{n \times 3}$, our model aims to generate a skeletal structure with joints $\mathbf{J} \in \mathbb{R}^{k \times 3}$ and connectivity $\mathbf{B} \in \mathbb{R}^{b \times 2}$.

Traditional learning methods for $\mathbf{J}$ (joint positions) and $\mathbf{B}$ (bone connections) typically adopt a multi-stage regression approach~\cite{xu2020rignet, ma2023tarig}, where $\mathbf{J}$ is learned first, followed by the estimation of parent-child relationships, and then $\mathbf{B}$ is inferred using greedy algorithms such as Minimum Spanning Tree (MST). However, this multi-stage learning paradigm suffers from several limitations: (1) Error accumulation occurs across stages, leading to inaccuracies in connectivity prediction.
(2) Greedy algorithms like MST struggle to generate satisfactory results for complex skeletal structures.
(3) Regression-based methods lack generalization capability when dealing with large-scale, multi-category datasets.

To overcome these challenges, we jointly consider both $\mathbf{J}$ and $\mathbf{B}$ and obtain a tree-structured skeleton representation $\mathbf{T}$ through them as:
\begin{equation}\label{eqn:1}
\begin{aligned}
& \mathbf{T}_{1:k} = [\mathbf{P}_{1}, \mathbf{J}_{1}, \mathbf{P}_{2}, \mathbf{J}_{2},...,\mathbf{P}_{k}, \mathbf{J}_{k}].
\end{aligned}
\end{equation}
where $\mathbf{J}_{i} \in \mathbb{R}^{3}$ represents the position of the $i^{th}$ joints and $\mathbf{P}_{i} \in \mathbb{R}^{3}$ represents the position of the parent joint of the $i^{th}$ joints. 
We then introduce an autoregressive model that predicts skeleton structures in a sequential manner, conditioned on mesh vertex inputs. Unlike traditional multi-stage regression approaches, our method jointly learns both joint positions and connectivity relationships, effectively mitigating error accumulation and improving generalization across diverse categories. 
This process can be formulated as:

\begin{equation}\label{eqn:2}
\begin{aligned}
& P(\mathbf{P}_{k},\mathbf{J}_{k}|\mathbf{T}_{1:k-1},\mathbf{V}) = P(\mathbf{P}_{k}|\mathbf{T}_{1:k-1},\mathbf{V}) \\
&\quad P(\mathbf{J}_{k}|\mathbf{P}_{k},\mathbf{T}_{1:k-1},\mathbf{V}).
\end{aligned}
\end{equation}
For the specific implementation of conditional generation, simply using a condition autoregressive model can lead to misalignment between the input point cloud and the skeletal joints. Inspired by~\cite{tang2024edgerunner}, we adopt an Autoregressive Auto-Encoder (ArAE) to facilitate autoregressive learning, followed by latent diffusion for conditional generation.

\subsection{Skeleton Generation}\label{sec:skeleton}

\noindent\textbf{Skeleton tokenization: }Next, we detail the tokenization process for skeletal joints. We represent the skeleton using a tree structure $\mathbf{T}$, where each joint is a node. To ensure a consistent numerical range, we first normalize the joint positions $\mathbf{J} \in \mathbb{R}^{k \times 3}$, mapping their coordinates to the range $[-1,1]$. For each joint $\mathbf{J}_{i} \in \mathbb{R}^{3}$, we discretize its x, y, z coordinates, yielding three tokens. Similarly, we apply the same discretization process to its parent joint $\mathbf{P}_{i} \in \mathbb{R}^{3}$, resulting in an additional three tokens. As is shown in Fig.~\ref{fig:pipeline}, the gray tokens means the parent tokens, and the yellow ones indicate the joints. In addition, we assume that the first three tokens represent the root node of the tree structure.
Ultimately, this formulation produces a total of $6k$ skeleton tokens $\mathbf{O}$, which serve as the label for our autoregressive model.

\noindent\textbf{Auto-regressive Auto-encoder: }We adopt an autoregressive auto-encoder model to establish a mapping between the joints $\mathbf{J}$ and the corresponding skeletal tokens as shown in Fig~\ref{fig:pipeline}(a). 
The latent embedding obtained from the auto-encoder serves as the conditioning input for the latent diffusion model during conditional generation.

Specifically, we first encode the padded joints $\mathbf{J}$ using a joints encoder $\mathbf{F_{j}}$ to obtain the corresponding latent embedding $\mathbf{L}$. 
\begin{equation}\label{eqn:3}
\begin{aligned}
& \mathbf{L} = \mathbf{F_{j}}(\mathbf{Q}, Pos(\mathbf{J})),
\end{aligned}
\end{equation}
where $\mathbf{Q}$ is a learnable query embedding that aims to compress the input data, and $Pos$ is a frequency embedding function~\cite{tang2024edgerunner}.
We choose cross attention layer as the joints encoder here.
This latent embedding $\mathbf{L}$ is then used as the conditioning input for the autoregressive model $\mathbf{F_{a}}$. 
To initiate the token prediction process, we append a BOS token after the latent embedding and employ a next token prediction strategy to sequentially generate the skeletal token sequence $\hat{\mathbf{O}}$.
\begin{equation}\label{eqn:4}
\begin{aligned}
& \hat{\mathbf{O}}_{i} = \mathbf{F_{a}}(\hat{\mathbf{O}}_{1:i-1}, \mathbf{L}),
\end{aligned}
\end{equation}
where $\hat{\mathbf{O}}_{i}$ is the predicted $i^{th}$ skeletal token. Our model is trained end-to-end using the cross-entropy ($\mathcal{L}_{ce}$):
\begin{equation}\label{eqn:5}
\begin{aligned}
& \mathcal{L}_{ce} = CE(\hat{\mathbf{O}}[:-1], \mathbf{O}[1:]).
\end{aligned}
\end{equation}

\noindent\textbf{Mesh Condition Generation: }Our goal is to generate the corresponding skeleton from the given mesh condition. 
Since the trained ArAE model can decode skeleton tokens from the latent embedding obtained via the joints encoder, we employ a conditioned latent diffusion model to learn this latent embedding as shown in Fig~\ref{fig:pipeline}(b).

Specifically, for a given mesh input, we first sample 1,024 surface points. These points are then processed by a pre-trained point cloud encoder~\cite{qi2024shapellm}, which extracts the corresponding feature representation. 
This feature is subsequently refined using a cross-attention layer to obtain the denoised feature.

During training, both the joints encoder and the point cloud encoder are kept fixed. We adopt the DDPM framework~\cite{ho2020denoising} and use the MSE loss to train the latent diffusion model. 
At inference time, for any given input mesh, the point cloud encoder extracts the corresponding feature, which is then fed into the latent diffusion model to predict the latent embedding. This predicted latent is subsequently decoded by the autoregressive model to generate the final skeleton tokens.

\noindent\textbf{Skeleton Detokenization: }
Through the predefined skeleton tokenization, every three tokens can be viewed as a union, indicating the coordinates of the predicted points. We then view the odd unions as the parents for the even unions, which represent the estimated joints. For each parent union, we use the nearest neighbor search to find the corresponding point in joints. Thanks to the autocorrelation pattern of the autoregressive model, our tokenization process can generate correct and solid parent-children relationships without further learning networks or complicated postprocessing.
After obtaining high-quality skeleton data, we employ the state-of-the-art skinning estimation method GeoVoxel (GVB)~\cite{dionne2013geodesic}, to achieve reliable and accurate skinning results.

%% file: figs/pipeline.tex
\begin{figure*}[h!]
  \begin{center}
\includegraphics[width=17.5cm]{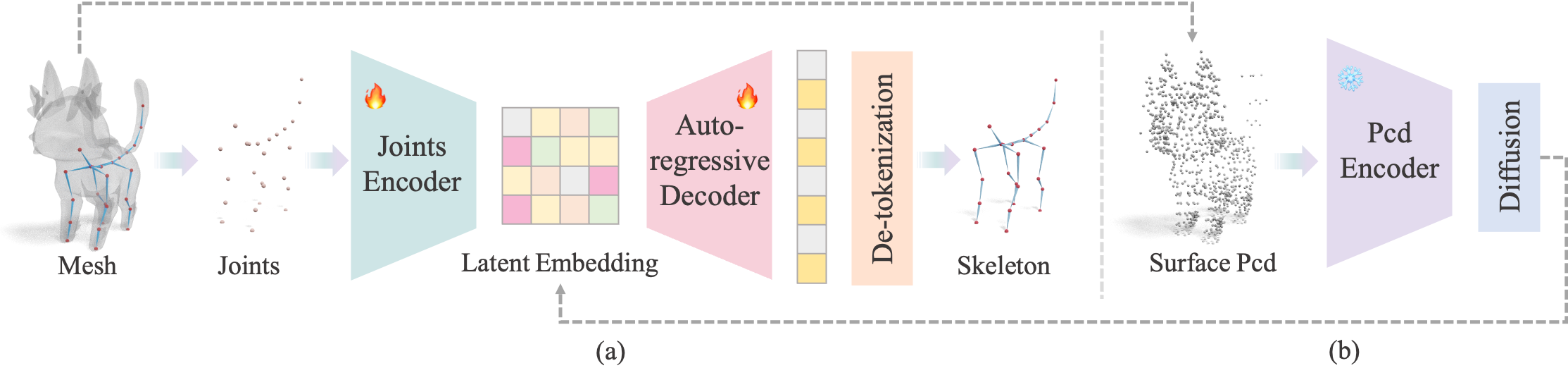}
  \end{center}
\caption{The overall pipeline of our framework. (a) An autoregressive auto-encoder model to establish the latent embedding for the skeleton. (b) A conditioned latent diffusion model to align the skeleton with the mesh through latent features. See Sec.\ref{sec:skeleton} for more details.}
\label{fig:pipeline}
\end{figure*}

%% file: sec/5_exp.tex
\section{Experimental Results}

\subsection{Implementation Details}
Our training process consists of two stages, where both the auto-regressive auto-encoder model and the latent diffusion model conditioned on point clouds require approximately one day of training on 8 A100 GPUs. 
The batch size is set to 128.
We split the data from each category into training and test sets with a 20:1 ratio and then merge them to form the final training and test datasets.
To ensure class balance during training, we randomly select 20\% of the data from the character category in each epoch.
To enhance the model’s robustness to input variations, we incorporate online pose augmentation in both training stages. Specifically, during training, we randomly deform joint positions while preserving connectivity relationships and use ground truth skinning weights to deform the corresponding mesh. 
The ablation study in Sec.~\ref{sec:abl} validates the effectiveness of this data augmentation strategy.

\subsection{Metircs and Baselines}

\noindent\textbf{Metircs: }To assess the accuracy of the predicted skeletons maps, we employ the same evaluation metrics as RigNet~\cite{xu2020rignet}.
For skeleton evaluation, we utilize CD-J2J (Chamfer Distance between joints), CD-J2B (Chamfer Distance between joints and bones), CD-B2B (Chamfer Distance between bones), IoU (Intersection over Union), as well as Precision \& Recall.

\noindent\textbf{Baselines: }We compare our method with a classic optimization method and a state-of-the-art learning-based method: (1) Pinocchio~\cite{baran2007automatic} (2) RigNet~\cite{xu2020rignet}.
To meet the input requirements of Pinocchio, we preprocess all input meshes by applying a watertight transformation to ensure they are manifold.
For a fair comparison, we trained and evaluated RigNet on the same dataset as our method.

\subsection{Evaluation for Skeleton Prediction}
\noindent\textbf{Quantitative result:} 
Tab.~\ref{table:joints} presents the quantitative comparison in joint estimation. Our results significantly outperform Pinocchio and RigNet across all metrics.
Specifically, the IoU metric, which measures the quality of joint estimation shows a 13.2\% improvement, while the CD-B2B metric, which evaluates the accuracy of bone estimation, shows a 41.7\% improvement. 
Compared to using MST for connectivity estimation, our autoregressive model can simultaneously learn both accurate joint positions and connectivity relationships.

\input{tabs/joints}

\noindent\textbf{Qualitative result:} 
For qualitative evaluation, we select examples featuring a variety of categories, as shown in Fig.~\ref{fig:baselines}.
Pinocchio can only generate template skeletons for humanoid and animal models and is unable to produce skeletons for other object categories.
RigNet lacks generalization capability for complex shapes, failing to produce accurate joint positions and reasonable connectivity. As shown in Fig.~\ref{fig:baselines} (second row, first column), even when it achieves relatively accurate joint positions, its multi-stage training approach leads to cumulative errors, resulting in incorrect connectivity predictions. This further demonstrates the advantage of our autoregressive model, which simultaneously predicts both joint positions and connectivity.
Moreover, RigNet struggles to align with meshes when handling inputs with complex poses, as illustrated in Fig.~\ref{fig:baselines} (first row, second column). 
In contrast, our method mitigates this issue through online pose augmentation, ensuring better alignment.

\input{figs/baselines}

\input{figs/morecases}

\subsection{Ablation study}\label{sec:abl}
\noindent\textbf{Conditional generation:} 
For our model, the key process involves a two-stage training approach: Stage one reconstructs the skeleton using an auto-regressive auto-encoder, and stage two uses latent diffusion for the conditioning input to learn the latent embedding.
In the ablation study, we first evaluate condition generation by employing a point cloud diffusion model to predict joint coordinates. 
Following RigNet~\cite{xu2020rignet}, we then estimate connectivity using MST. 
As shown in Tab.~\ref{table:joints_abl} (Only Diffusion Model), directly generating joint positions through diffusion leads to inaccurate placements, while MST-based connectivity estimation suffers from reduced accuracy, resulting in a drop of 53.2\% in CD-B2B.
Another ablation experiment replaces the conditioning generation process with a GPT-based approach for skeleton generation. 
As shown in Tab.~\ref{table:joints_abl} (Only GPT Model), this method suffers from an alignment issue between the generated skeleton and the conditioning point cloud. 
The conditioning accuracy is lower, leading to an 11.8\% drop in precision and a 25.4\% drop in CD-J2J compared to our approach. 
These results further validate the effectiveness of our model design.

\noindent\textbf{Augmentation: }In our training, a key data augmentation technique involves randomly altering the input pose online. 
In the ablation study, we compare the performance without pose augmentation. 
As shown in Tab.~\ref{table:joints_abl} (Ours w/o pose aug.), the CD-J2J score decreases 3.9\%, indicating that pose augmentation significantly improves the accuracy of joint position prediction.

\input{tabs/joints_abl}

%% file: tabs/joints.tex
\begin{table}[htbp]
\centering
\setlength{\tabcolsep}{4pt} 
\resizebox{\columnwidth}{!}{ 
\begin{tabular}{ccccccc}
\hline
\rowcolor[HTML]{FFFFFF} 
& IoU $\uparrow$             & Prec. $\uparrow$           & Rec. $\uparrow$            & CD-J2J $\downarrow$          & CD-J2B $\downarrow$          & CD-B2B $\downarrow$          \\ \hline

\rowcolor[HTML]{FFFFFF} 
Pinocchio      & 36.47\%         & 39.68\%         & 38.43\%     & 8.45\%           & 7.55\%           & 6.78\%                \\
\rowcolor[HTML]{FFFFFF} 
RigNet      & 61.35\%         & 60.64\%         & 67.93\%     & 6.44\%           & 5.85\%           & 5.06\%                \\
\rowcolor[HTML]{E8E8E8} 
Ours                         & \textbf{70.68\%} & \textbf{69.84\%} & \textbf{71.94\%} & \textbf{3.88\%} & \textbf{3.15\%} & \textbf{2.95\%} \\ \hline
\end{tabular}
}
\caption{
Joint prediction results on the test set.
}\label{table:joints}
\end{table}

%% file: figs/baselines.tex
\begin{figure*}[h!]
  \begin{center}
\includegraphics[width=17.5cm]{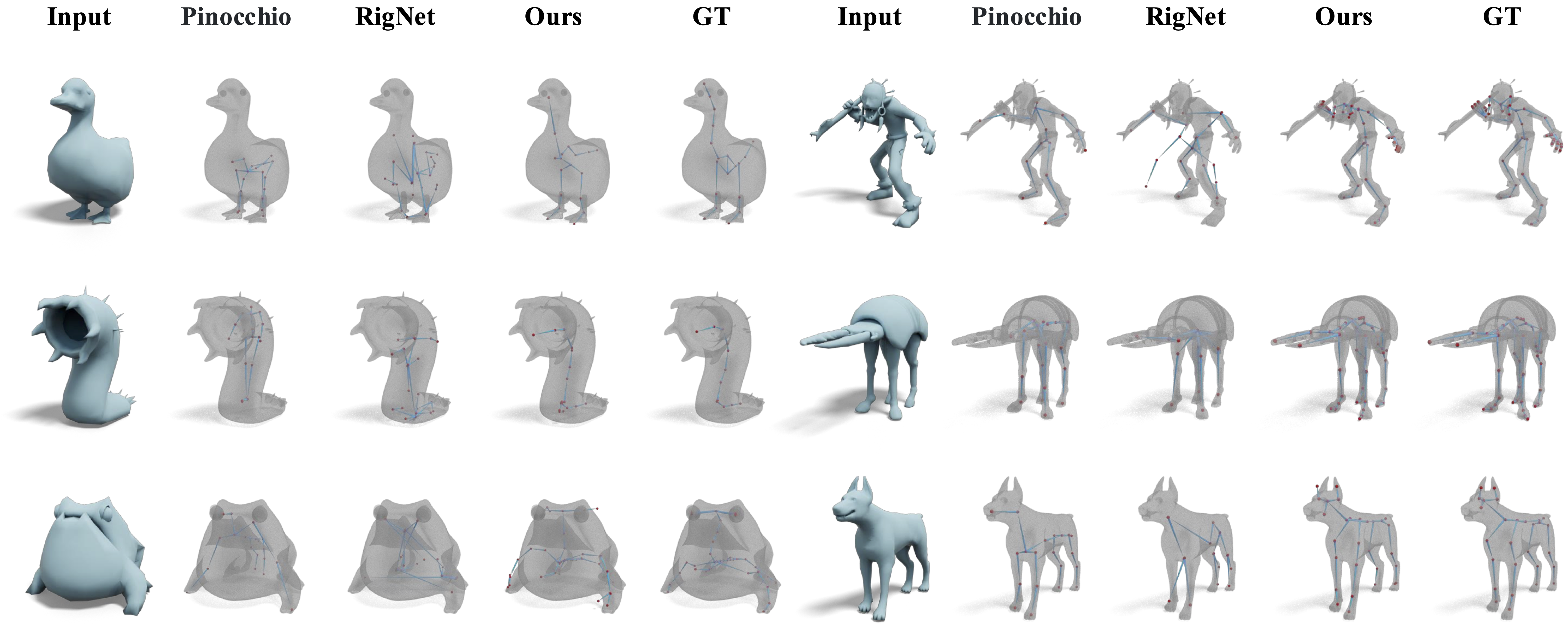}
  \end{center}
\caption{Comparison of skeleton generation results on $\dn$. Our method can generate reasonable skeleton results for diverse object categories and inputs with complex poses.}
\label{fig:baselines}
\end{figure*}

%% file: figs/morecases.tex
\begin{figure*}[h!]
  \begin{center}
\includegraphics[width=17.5cm]{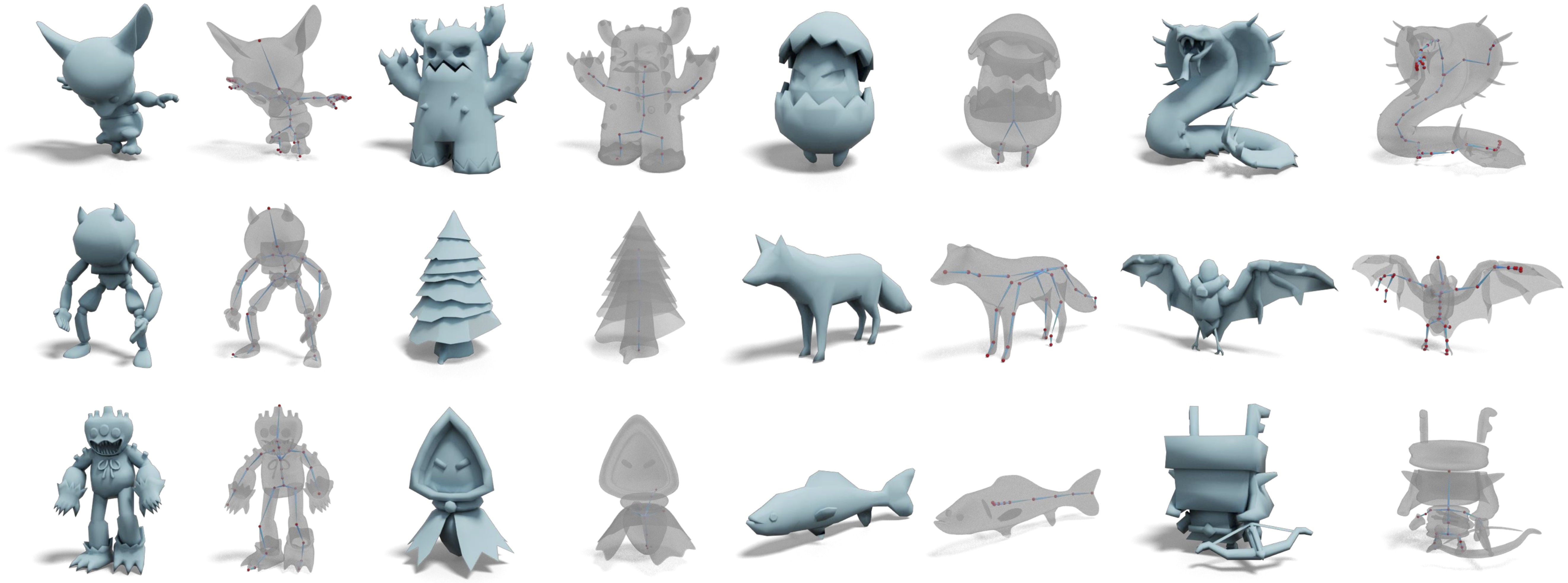}
  \end{center}
\caption{We present additional qualitative results of skeleton generation. Our model is capable of producing reasonable skeletal structures for inputs with diverse categories and varying poses.}
\label{fig:morecases}
\end{figure*}

%% file: tabs/joints_abl.tex
\begin{table}[!t]
\centering
\setlength{\tabcolsep}{4pt} 
\resizebox{\columnwidth}{!}{ 
\begin{tabular}{ccccccc}
\hline
\rowcolor[HTML]{FFFFFF} 
                       & IoU $\uparrow$             & Prec. $\uparrow$           & Rec. $\uparrow$            & CD-J2J $\downarrow$          & CD-J2B $\downarrow$          & CD-B2B $\downarrow$          \\ \hline

\rowcolor[HTML]{FFFFFF} 
Only Diffusion Model         &  56.68\%                &  55.90\%                &  58.28\%                  &  6.98\%               & 6.38\%                &  6.31\%               \\
Only GPT Model &  62.37\%                &  61.73\%                &  63.23\%                 & 5.20\%                &  5.36\%               & 4.91\%                \\
\rowcolor[HTML]{FFFFFF} 
Ours w/o pose aug.           &  66.79\%                &  68.17\%                &    67.89\%              &  4.04\%               & 3.51\%                &  3.20\%               \\
\rowcolor[HTML]{E8E8E8} 
Ours                         & \textbf{70.68\%} & \textbf{69.84\%} & \textbf{71.94\%} & \textbf{3.88\%} & \textbf{3.15\%} & \textbf{2.95\%} \\ \hline
\end{tabular}
}
\caption{Ablation study on joint estimation. w/o pose aug. denotes training without online pose augmentation.}\label{table:joints_abl}
\end{table}

%% file: sec/6_app.tex
\section{Application}
After obtaining an accurate skeleton and reasonable skinning results, animating heterogeneous skeletons still typically requires manual effort, which is both time-consuming and labor-intensive. 
To fully leverage our skeleton prediction model, we explore an automated motion transfer solution as a practical application.

Given a target mesh $\mathbf{M}_{T}$ with vertices $\mathbf{V}_{T}\in \mathbb{R}^{N_{T}\times k}$, our model can generate a skeletal structure with joints $\mathbf{J}_{T}\in \mathbb{R}^{K_{T}\times k}$ and connectivity. We can also obtain initial skinning weights $\mathbf{Skin}_{T}\in \mathbb{R}^{N_{T}\times K_{T}}$ using previous methods~\cite{dionne2013geodesic}. Then, our goal is to create motions based on the rigging information. For motion transfer, recent methods~\cite{zhang2024magicpose4d,xu2022morig,maheshwari2023transfer4d} first construct the skeletal structure for the source mesh sequences and transfer the skeleton to the target mesh, which highly rely on sequential features or template structures. In contrast, we view the source sequence as a guide and learn the transformation of each joint based on the skeleton of the target mesh. We use DT4D~\cite{li20214dcomplete} as the reference motion sequence.

Assume that the source sequence contains several frames and the corresponding mesh $\mathbf{M}_{S}^{(t)}$ with vertices $\mathbf{V}_{S}^{(t)}\in \mathbb{R}^{N_{S}^{(t)}\times k}$ at frame t. In order to learn the motions of the source sequence, we first transfer the skeleton of $\mathbf{M}_{T}$ to $\mathbf{M}_{S}^{(t)}$ through the correspondence shape matching. Several sophisticated methods~\cite{sun2024srif, eisenberger2020smooth, jiang2023non, sun2023spatially} have been proposed to solve non-rigid shape matching and registration. Thus, through shape mapping $\mathbf{Map}_{S^{(t)}, T}\in \mathbb{R}^{N_{S}^{(t)}\times N_{T}}$, we can obtain the source skinning $\mathbf{Skin}_{S^{(t)}}\in \mathbb{R}^{N_{S^{(t)}}\times K_{T}}$. After that, the joints of source mesh $\mathbf{J}_{S^{(t)}}\in \mathbb{R}^{K_{T}\times k}$ can be calculated through arithmetic mean based on the transferred skinning and vertices $\mathbf{V}_{S}^{(t)}$. We then use a lightweight optimization network to learn the rotation and translation for different joints according to the meshes from the source sequence. Note that the joints for different frames are adaptive; the skeleton motion can be further enhanced by the various postures. Since the skeleton for the target and source mesh are consistent, we can directly apply the optimized transformation on the target mesh. As is shown in Fig.~\ref{fig:transfer}, with the high-quality skeleton provided by our base model, we can realize motion transfer in several categories, varying from animals to human.

\input{figs/transfer}

%% file: figs/transfer.tex
\begin{figure}[h!]
  \begin{center}
\includegraphics[width=\linewidth]{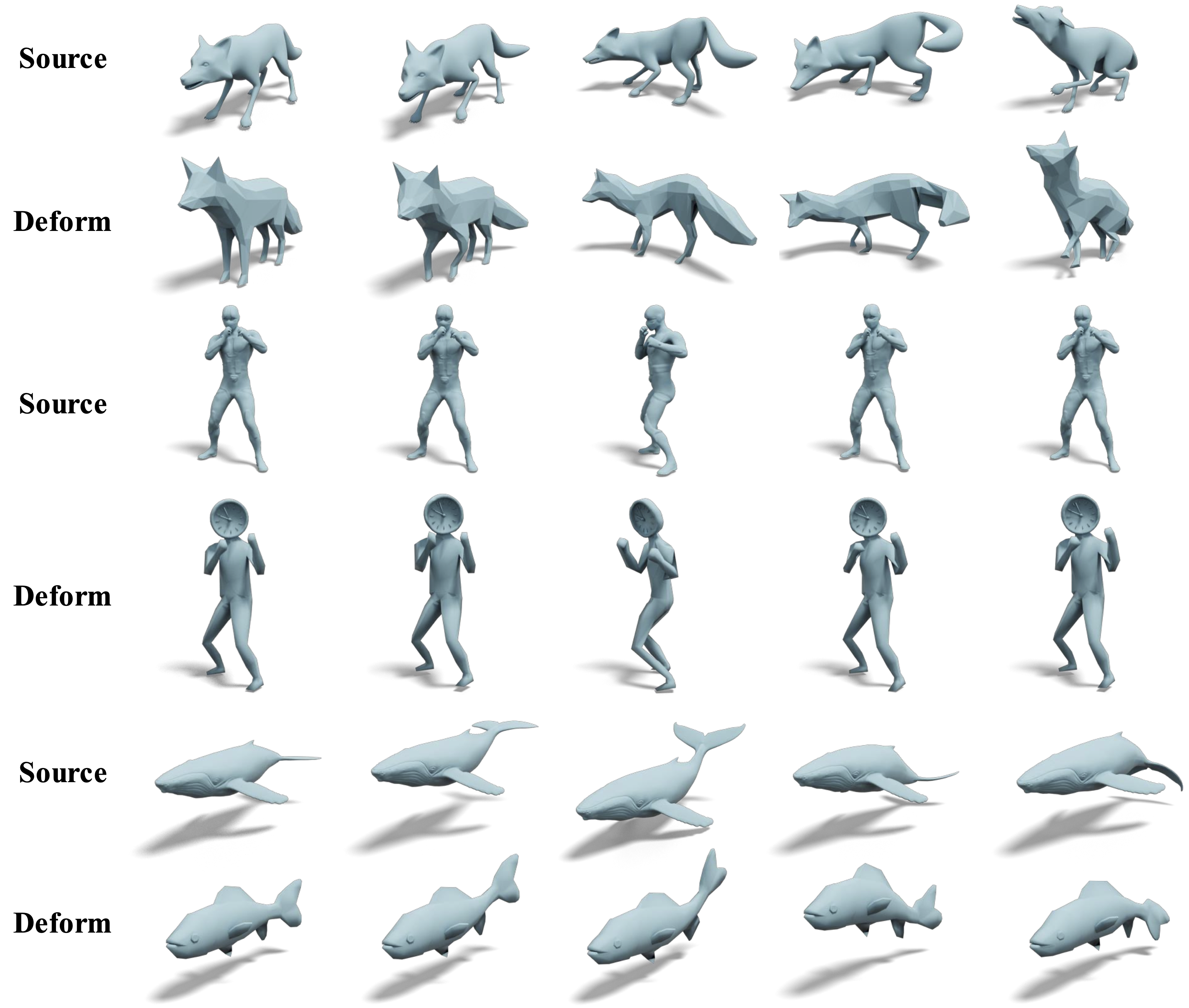}
  \end{center}
\caption{The visualization of the motion transfer results guided by DT4D~\cite{li20214dcomplete}. Odd rows indicate the meshes from the source sequences and even rows show the transferred motions on the target mesh with well-defined skeletons.}
\label{fig:transfer}
\end{figure}

%% file: sec/7_conclusion.tex
\section{Conclusion, Limitation, and Future Work}
In this paper, we present \texttt{ARMO}, a novel rigging framework designed to predict accurate skeletal structures for 3D models. 
To support our approach, we introduce $\dn$, the first large-scale rigging dataset, featuring 79,499 models with comprehensive skeleton and skinning information. 
Our dataset expands the scope of rigging research by incorporating diverse shape categories, styles, and poses, moving beyond the constraints of traditional benchmarks.
Our proposed method addresses key limitations in existing rigging algorithms. 
By employing an autoregressive model, we achieve simultaneous prediction of joint positions and connectivity relationships, mitigating error accumulation inherent in multi-stage methods. 
Additionally, our mesh-conditioned latent diffusion model further enhances prediction accuracy and generalization.
Extensive experiments demonstrate that \texttt{ARMO} outperforms existing methods on the $\dn$ dataset in skeleton prediction. We believe that our dataset and method will serve as a strong foundation for future advancements in 3D rigging, pose estimation, and animation synthesis, paving the way for more versatile and dynamic 3D content generation.

We also identify the following limitations, which lead to future work directions: 
1) The node density of our generated skeleton is fixed and cannot be adjusted based on user preferences. Incorporating more versatile data sources or introducing a node density embedding within the autoregressive model could offer greater flexibility in controlling node density;
2) Our model struggles to produce fully consistent skeleton results for sequence data, potentially due to limited data augmentation and pose-awareness during training. Exploring these aspects will be an interesting direction for future research.